\documentclass[conference]{IEEEtran}
\IEEEoverridecommandlockouts
\usepackage{cite}
\usepackage{amsmath,amssymb,amsfonts}
\usepackage{algorithmic}
\usepackage{graphicx}
\usepackage{textcomp}
\usepackage{xcolor}
\usepackage{url}
\usepackage{hhline}
\def\BibTeX{{\rm B\kern-.05em{\sc i\kern-.025em b}\kern-.08em
    T\kern-.1667em\lower.7ex\hbox{E}\kern-.125emX}}
\begin{document}

\title{Road Damage Detection and Classification with Detectron2 and Faster R-CNN}

\author{
\IEEEauthorblockN{Vung Pham}
\IEEEauthorblockA{\textit{Computer Science Department} \\
\textit{Texas Tech University}\\
Lubbock, USA \\
vung.pham@ttu.edu}
\and
\IEEEauthorblockN{Chau Pham}
\IEEEauthorblockA{\textit{Computer Science Department} \\
\textit{Texas Tech University}\\
Lubbock, USA \\
chaupham@ttu.edu}
\and
\IEEEauthorblockN{Tommy Dang}
\IEEEauthorblockA{\textit{Computer Science Department} \\
\textit{Texas Tech University}\\
Lubbock, USA \\
tommy.dang@ttu.edu}

}

\maketitle

\begin{abstract}
The road is vital for many aspects of life, and road maintenance is crucial for human safety. One of the critical tasks to allow timely repair of road damages is to quickly and efficiently detect and classify them. This work details the strategies and experiments evaluated for these tasks. Specifically, we evaluate Detectron2's implementation of Faster R-CNN using different base models and configurations. We also experiment with these approaches using the Global Road Damage Detection Challenge 2020, A Track in the IEEE Big Data 2020 Big Data Cup Challenge dataset. The results show that the X101-FPN base model for Faster R-CNN with Detectron2's default configurations are efficient and general enough to be transferable to different countries in this challenge. This approach results in F1 scores of 51.0\% and 51.4\% for the test1 and test2 sets of the challenge, respectively. Though the visualizations show good prediction results, the F1 scores are low. Therefore, we also evaluate the prediction results against the existing annotations and discover some discrepancies. Thus, we also suggest strategies to improve the labeling process for this dataset.

\end{abstract}


\section{Introduction}
The road is crucial in different aspects of life, from economic development and social benefits to safety. Therefore, road maintenance is vital for all countries in the world. One of the road maintenance tasks is to accurately detect damages and then to devote efficient repairs in a timely manner. However, for most countries, road crack detection and classification are currently based on human manual works or expensive sensors. Therefore, automatic detection and classification of road damage types are getting popular recently. Also, deep learning, with its recent advancements, is gaining traction and has state-of-the-art results in various computer vision tasks~\cite{scagcnn}. Therefore, many works in the literature use deep learning approaches to detect and classify road damages. 

Using deep learning for road damage detection and classification often involves three sub-tasks: 1) collecting image data, 2) creating labels for the data, and 3) building deep learning models from the labeled data. While collecting data can be done efficiently using mobile devices with GPS and camera~\cite{maeda2018road}, the labeling process takes time, and the detection/classification results are still limited. Also, the models learned from the data coming from one country often are not generalized enough to be transferable to different countries~\cite{arya2020transfer}. Additionally, providing bounding boxes and labels for road damages is error-prone and demands a massive amount of human labor to have accurate results. 

Therefore, this work explores state-of-the-art object detection methods to find general road damage detection and classification model that is usable for different territories. We then apply the selected approaches to the Global Road Damage Detection Challenge 2020, A Track in the IEEE Big Data 2020 Big Data Cup Challenge~\cite{bigdata2020cup} dataset. Furthermore, we also evaluate the quality of the currently labeled data and propose a strategy to generate more labeled data. Thus, our contributions are:
\begin{itemize}
    \item Exploring current state-of-the-art object detection methods and their applicability to road damage detection and classification tasks.

    \item Experimenting with these approaches using the Global Road Damage Detection Challenge 2020 dataset to find one single model that is efficient and can be transferable to different territories.
    
    \item Visualizing the prediction results and qualitatively evaluating the existing annotations follows by giving suggestions to improve the labeling process for this dataset. 
\end{itemize}

\section{Dataset and evaluation method}
The dataset used in this work is from the Global Road Damage Detection Challenge 2020 \cite{bigdata2020cup}. This dataset consists of one train set (\textit{train}) and two test sets (\textit{test1} and \textit{test2}).  Specifically, the training and testing sets contain road damage images, bounding boxes, and damage types (for training set) from three countries (i.e., Czech, India, and Japan). Furthermore, for this dataset, there are four types of labelled road damage types namely Longitudinal Crack, Transverse Crack,  Aligator Crack, and Pothole (labelled \textit{D00}, \textit{D10}, \textit{D20}, and \textit{D40}, respectively). For this section's brevity, we defer further details about the number of images and damage type distributions in the individual countries and the whole dataset to Section~\ref{sec:dataexploration}.

For this challenge, the evaluation method is based on the \textit{F1} score defined to balance the precision (\textit{p}) and recall (\textit{r}). These are defined as:
\begin{equation*}
    p = \frac{C_d}{P_d},\text{   } r = \frac{C_d}{A_d},\text{   } F1 = \frac{2pr}{p + r}
\end{equation*}
where $C_d$, $P_d$, and $A_d$ are the numbers of correctly predicted damages, the predicted damages, and all the ground-truth damages from the evaluating set, respectively. Furthermore, the definition of correctly predicted damage has two criteria. They are 1) the predicted bounding box must match, and 2) the predicted label is correct. The latter is obvious, and the former is defined by the Intersection over Union (\textit{IoU}) score, which is defined as follows:
\begin{equation*}
    IoU = \frac{area(P_b\cap G_b)}{area(P_b\cup G_b)}
\end{equation*}
where $P_b$ and $G_b$ are a predicted box and a ground-truth box, respectively. Also, $area(P_b \cap G_b)$ and $area(P_b \cup G_b)$ means the areas of the intersection and the union between the two boxes, correspondingly. In this case, if $IoU >= 0.5$, then it is a match, and it is not otherwise.

\section{Related work}
Object detection using deep neural networks is an emerging field, and it is not the purpose of this work to review all the recent results in this field. Also, object detection techniques are continually evolving, and comparisons might be outdated quickly. Instead, we briefly survey families of current state-of-the-art object detection methods in general and methods related to road damage detection and classification specifically.

\subsection{Deep learning based object detection}
Deep learning-based object detection is gaining initial success, and there are many works in the literature regarding this. We refer interested readers to~\cite{tong2020recent} for a good survey of these methods. Recent techniques are generally related to two prominent families, namely Region-Based Convolutional Neural Networks (R-CNNs) and You Only Look Once (YOLO).

Ross Girshick et al.~\cite{girshick2014rich} propose R-CNNs approach with three main modules. The first one is a region proposal module that generates candidate regions (bounding boxes) using computer vision techniques. The second one is the feature extraction module. This second module uses convolutional neural networks to extract the features from the candidate regions. Finally, the last module is a classifier that predicts the classes of the proposed candidates using the extracted features. 

R-CNNs takes a long time to train because training is done in multiple stages. Besides training, the prediction stage is also slow. Therefore, Girshick proposes another model called Fast R-CNN~\cite{girshick2015fast} to tackle these issues. Fast R-CNN is trained as a single model instead of three separate modules. This architecture takes the images and proposes candidate regions, then passes them through a popular, pre-trained image classification model (e.g., ResNet~\cite{he2016deep}, VGG-16~\cite{vgg16}) to extract features from the candidates. The extracted features then undergo a Region of Interest (RoI) pooling layer, followed by two fully connected layers. Finally, there are two other fully connected heads for bounding box regression and label classification purposes.

Though Fast R-CNN improves the training and predicting time, it still needs the region proposal as the inputs. In other words, the region proposals for each image still needs to be done separately (e.g., using image processing techniques). Therefore, Ren et al.,~\cite{ren2015faster} propose Faster R-CNN to tackle this issue. Its main improvement is the ability to incorporate region proposals as a part of the final model using Region Proposal Network (RPN). In other words, there are two smaller networks in this architecture. The first one is a Region Proposal Network (RPN), and the second one is the Fast R-CNN. These two sub-networks are trained simultaneously, though for two different tasks: 1) region proposals and 2) bounding box classification and regression. These strategies help to improve the training and object detection time and accuracy.

Another famous family of object detection is YOLO with different versions such as YOLO~\cite{redmon2016you}, YOLOv2~\cite{redmon2017yolo9000}, YOLOv3~\cite{redmon2018yolov3}, and YOLOv4~\cite{bochkovskiy2020yolov4}. Different YOLO versions may differ in terms of architectures and techniques used. However, generally, it involves a single neural network with the input being the images and ground-truth boxes/segments and labels (while training). The outputs are the bounding boxes and corresponding labels of the detected objects from the image. Specifically, it divides an image into a grid of cells. Feature extracted from each cell is used to predict objects with centers of the bounding boxes that fall into the cell. The advantage of this method is that it is faster to train and predict. However, the benefit comes with a slightly lower accuracy compared to Faster R-CNNs.

While YOLO mainly has its advantage for speed and Faster R-CNN is better at accuracy, Single Short Detection (SSD)~\cite{liu2016ssd} allows a better balance between speed and accuracy. SSD runs a convolutional neural network on input image only one time and computes a feature map. It also uses anchor boxes at different sizes and aspects as Faster R-CNN. Regarding bounding box sizes, SSD predicts them at different convolutional layers. The reason is that convolutional layers have different receptive fields of the inputs. In other words, the deeper a convolutional layer is, the larger its receptive field will be. Thus, the deeper convolutional layer features are used to predict larger bounding boxes and vice versa.

\subsection{Deep learning based road damage detection}
\label{sec:deeplearningforrdd}

Deep learning is gaining success in various areas, and road damage detection and classification is no exception. Various works in the literature use deep learning for these tasks. With the help of GPS- and camera-enabled mobile devices, it is now relatively easy to collect road images for this purpose. For instance, Maeda et al.,~\cite{maeda2018road} propose using a smartphone placed on a car's dashboard to collect pictures of road cracks, label them, and make them available for the public. Based on this dataset, Yanbo Wang et al.,~\cite{wang2018deep} propose to use Faster R-CNN and SSD with pre-trained ResNet and VGG as bases to tackle the damage detection and classification tasks. Also, they aggregate up to 14 Faster R-CNN models and 2 SSD models to improve the detection result. Similarly, Wenzhe Wang et al.,~\cite{wang2018road} also propose to use Faster R-CNN with data exploration step to adjust appropriate hyperparameters such as anchor boxes and ratios. Furthermore, they use different augmentation types (such as contrast transformation, brightness adjustment, and Gaussian blur) to improve their results.

On the other hand,~\cite{alfarrarjeh2018deep} uses YOLOv3 with \textit{darknet53} as the base model to tackle these tasks. They also experiment with two augmentation strategies. The first strategy is to generate more images for damage types with lower number occurrences using brightening or gray-scale, and the second strategy is to use cropping. However, in general, augmentations do not help. Additionally, Kluger et al., \cite{kluger2018region} utilize Faster R-CNN, RetinaNet~\cite{lin2017focal}, and Convolutional Neural Network combined with Random Forest to solve the tasks. The use of Random Forest is due to the assumption that it helps in case there are few samples in the training data~\cite{reinders2018object}. Furthermore, they propose (without validating the impacts of the proposal) to use Cycle-Consistent Adversarial Networks (CycleGAN)~\cite{zhu2017unpaired} for data augmentation. Their experiments also show that Faster R-CNN works best for road damage classification and detection.

In a recent study, Maeda et al.,~\cite{maedagenerative} use Generative Adversarial Nets (GAN)~\cite{goodfellow2014generative} to generate damages with a lower number of occurrences to improve the detection results for this specific type of damages. Specifically, they use Progressive Growing GAN (PG-GAN)~\cite{karras2017progressive} to artificially generate more damages of \textit{pothole} damage type and improve prediction results for this type of cracks in Japan. The reason is that Japan has a low number of \textit{pothole} damages. Furthermore, they also use Poisson blending~\cite{perez2003poisson} to place the generated damages to the existing images to make the artificial patches look more natural to its containers. In another work, Arya et al.,~\cite{arya2020transfer} suggest that the current techniques are not transferable from one country to another. Therefore, such a model is needed to save data collection, data labeling, and training time.

All in all, these works show that Faster R-CNN seems to be a useful technique for road damage detection and classification with a good trade-off between time and accuracy. These works also indicate that except when we only focus on a damage type with a small number of occurrences (i.e., \textit{pothole} damages in Japan), data augmentations (other than the obvious ones such as horizontal flipping and resizing) do not generally help. Furthermore, several of these works use ensembles to improve their prediction results. Though ensembles help improve the results, they also significantly increase training and prediction time and are not encouraged in this year's challenge.

\section{Methodology}

Our general methodology is that we start with the data exploration stage to understand the dataset. We then proceed by splitting the training dataset further into the training and evaluation sets. The validation enables us to evaluate the hyper-parameters for our architectures quantitatively. Regarding deep learning model architectures, we start with the commonly used model architectures for road damage detection and classification tasks. We then proceed with strategies to improve the base models, such as changing hyper-parameters, train data augmentations, and test time augmentations. It's worth noting that ensembles and individual models for individual countries would have better prediction results. However, it is restricted by the challenge that we should have a single-algorithm and single-model approach. Therefore, we do not attempt these directions.

\subsection{Data exploration and train/evaluation splits}
\label{sec:dataexploration}

\begin{figure}[!htb]
    \centering
    \includegraphics[width=\linewidth]{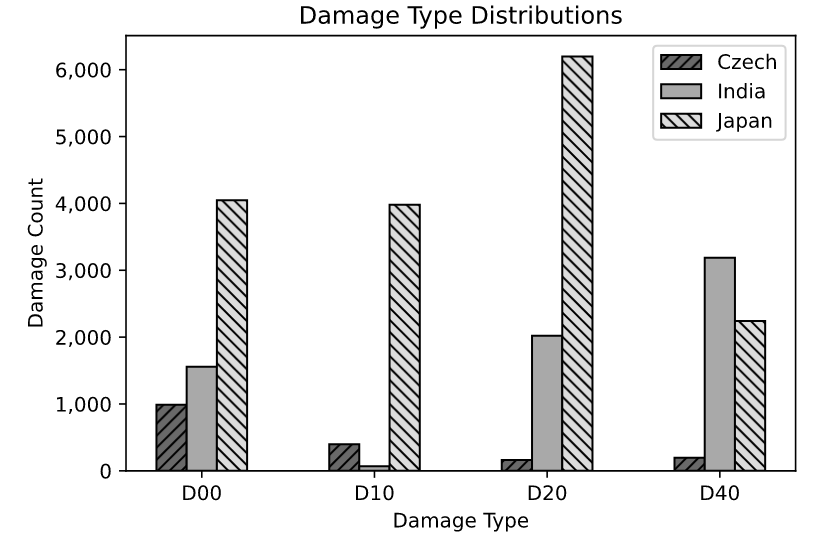}
    \caption{Damage type distributions over three countries (Czech, India, and Japan).}
    \label{fig:damagedistributions}
\end{figure}

This dataset consists of one training set (\textit{train}) and two test sets (\textit{test1} and \textit{test2}). The training set contains 21,041 images (2,829, 7,706, and 10,506 for Czech, India, and Japan, respectively). The two test sets contain 2,631 and 2,664 images, correspondingly. There are 34,702 ground-truth labels (bounding boxes and damage types) in the training set. Specifically, Figure \ref{fig:damagedistributions} shows the damage type distributions (of the four corresponding types) over the three countries. In general, Japan has higher numbers of images and damages, and the pothole damage type has the lowest number of occurrences. 

On the other hand, India has fewer images and damage labels. Also, \textit{D04} is the one with the highest number of occurrences, while there are only a few \textit{D01} damages in India. Finally, Czech has fewest number of images and damage samples (a total of 1,745 labels). It's worth noting that due to different numbers of images, different damage types distributions over different countries, it is difficult to have a transferable model for all three countries. In other words, having different models for different countries (with a sufficient number of images like Japan and India) should have better accuracy compared to a single model for all three countries.

\begin{figure}[!htb]
    \centering
    \includegraphics[width=\linewidth]{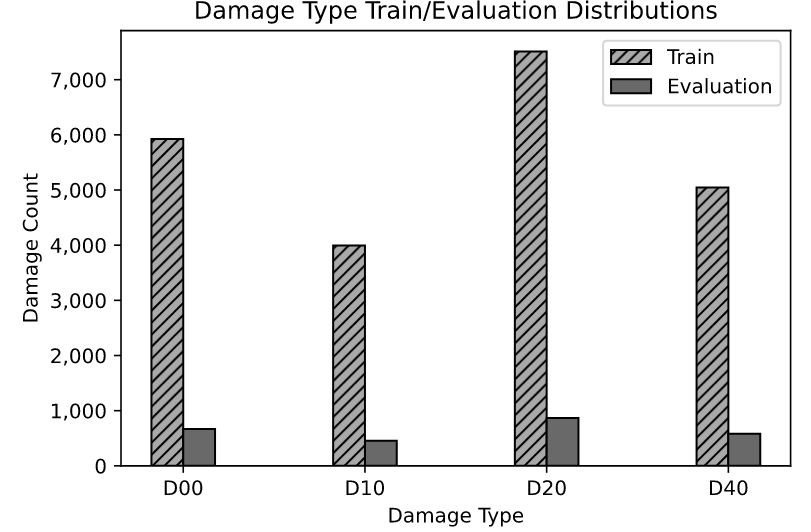}
    \caption{Damage type distributions of the training dataset after splitting it into training and evaluation sets.}
    \label{fig:trainevalsplits}
\end{figure}

We also split the training dataset into training and validation sets with the stratify as the origin of the damage types (i.e., country). Specifically, we keep 90\% of the images for training and 10\% for evaluation. This 10\% for the evaluation split is reasonable because it results in 1,221 images in this set. More than a thousand images are general enough to evaluate the performance of the learned models. This evaluation set is used to quantitatively evaluate our model's performance and the hyperparameter selection process during training (such as prediction score threshold and number of train iterations). Figure \ref{fig:trainevalsplits} shows the distributions of the damage types after splitting. Generally, though different countries have different damage type distributions, the combined distribution has a better balance among damage types. Also, using the origins of damage types as a stratified field seems sufficient since the train vs. evaluation distributions are somewhat similar. 

\subsection{Base models}

As discussed in Section \ref{sec:deeplearningforrdd}, most of the successful road damage detection and classification techniques trained using the Road Damage Detection dataset (version 2018) use the Faster R-CNN technique. Therefore, we first explore this approach to tackle this year's challenge. Instead of developing a Faster R-CNN model from scratch, we use Detectron2~\cite{wu2019detectron2} to speed up our development cycle. Detectron2 is Facebook AI Research's next-generation software system that implements state-of-the-art object detection algorithms.

It is also a common practice to use a base model pre-trained on a large image set (such as ImageNet~\cite{deng2009imagenet}) as the feature extractor part of the network. Detectron2 provides many such base-models~\cite{detectron2modelzoo}. However, for Faster R-CNN, two commonly used models are R101-FPN~\cite{r101fpn} and X101-FPN~\cite{x101fpn}. We select to explore these two pre-trained models because they have good Faster R-CNN box Average Precision (AP) compared to others. They are 42\% and 43\%, respectively. Though X101-FPN has better box AP on the ImageNet benchmark, it takes longer to train/predict and might be overfitting in some cases. That is the reason why we also explore R101-FPN.

There are many hyper-parameters to be tuned while training a Faster R-CNN model. Thus, it is nearly impossible in terms of time and computation resources to explore all the configurations. Therefore, we first stick with the default and obvious, common-sense configurations. Specifically, we set `cfg.SOLVER.REFERENCE\_WORLD\_SIZE' (number of GPUs) to 2, both `cfg.SOLVER.IMS\_PER\_BATCH' (images per batch) and `cfg.DATALOADER.NUM\_WORKERS' to 16, `cfg.SOLVER.BASE\_LR' (base learning rate) as 0.00025, and `cfg.MODEL.ROI\_HEADS.NUM\_CLASSES' (number of classes) is 4 (as correspond to four different types of damages). All other configurations are kept as default from Detectron2. Interested readers can refer to this page\footnote{\url{https://detectron2.readthedocs.io/modules/config.html}} for further details about these default configurations.

Although it is faster to train a Faster R-CNN using R101-FPN as the base model, it has a slightly lower accuracy than using X101-FPN as the base model (52.85\% vs. 54.25\% on the evaluation set at the \textit{test time prediction score thresholds} of 0.65 and 0.71, respectively). Specifically, it takes 0.82 seconds to train an iteration when using R101-FPN while it takes 1.67 seconds when using X101-FPN (using the specified configuration). The former also takes fewer iterations to converge than the latter (85,000 vs. 105,000). It takes time to do all the experiments with these models. Therefore, in the next sections, we only explore further experiments using the X101-FPN as the Faster R-CNN base.

\subsection{Data augmentations}

As discussed in Section \ref{sec:dataexploration}, though the training dataset has a good number of images and damage types, they are imbalanced among damage types and are distributed differently for different countries. Therefore, besides the apparent augmentation techniques (such as image resizing and horizontal flipping, called `default augmentations' hereafter), we also explore other augmentation techniques. Specifically, Maeda et al.,~\cite{maedagenerative} suggest that using GAN to generate synthetic damage types with fewer occurrences helps predict that specific type of damages (e.g., `pothole' type of damages). Generating images using GAN often involves three processes: 1) generating a patch for damage, 2) finding places in an existing image to blend the patch, and 3) making the synthetic patch as natural to (as if it really belongs to) the container image as possible.

The synthetic damage patch generation step using GAN takes a long time to train. Additionally, we can manually place the patches to an existing image~\cite{maedagenerative} or use another object detection model to detect the road areas and automatically position the artificial patches \cite{kluger2018region} to these areas. However, both of these approaches take time to complete. Therefore, in the following sections we detail our strategies to quickly evaluate the efficiency of this augmentation method before developing and training complicated models for these tasks, in case this direction is promising.

Instead of using GAN to generate the synthetic damage patches, we randomly extract and select existing patches using the given ground-truth boxes. We also apply some simple augmentations to the selected patches (such as horizontal flipping, slight rotating, or scaling) to increase the varieties of the damage patches. Furthermore, we also randomly sample the locations from all existing damages of the same type to position the artificial patch into an existing image. The reason is that we assume that similar damage types would appear in similar locations. This heuristic does not always hold, but it would be fast and easy to implement. 

\begin{figure}[!htb]
    \centering
    \includegraphics[width=\linewidth]{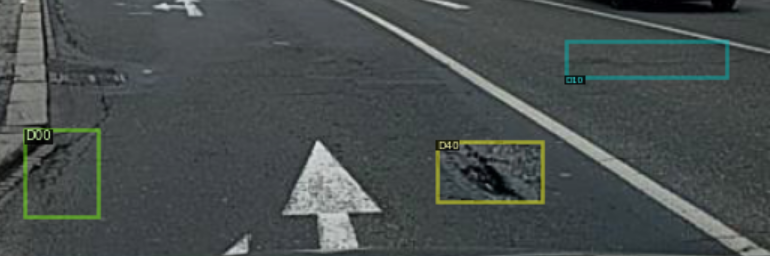}
    \caption{Damage augmentation: \textit{D00} (green-box) and \textit{D10} (blue box) are the real damages. \textit{D40} (yellow box) is the augmented one. This patch was randomly selected from existing \textit{D40} damages then slightly modified and placed into a position randomly sampled from all positions of this crack type.}
    \label{fig:augmentation}
\end{figure}

Furthermore, instead of using Poisson blending~\cite{perez2003poisson} to place the synthetic damage patches into an existing image, we use a fast color transfer technique \cite{colortransferring} to make the artificial patches look more natural to the placing image. Figure \ref{fig:augmentation} shows an example of the resulted image. \textit{D01} (green box) and \textit{D10} (blue box) damage boxes are the real ones, while \textit{D40} (yellow box) is the synthesized one. Though not all the artificial patches generated this way look natural or in the correct place, it helps to quickly generate augmented data and check if this augmentation strategy worth exploring.

Furthermore, in this case, we are not focusing on any specific type of damages but would like to balance the number of damages among the four damage types per country. Thus, we set different augmentation probabilities for different damage types in each country. Specifically, based on data exploration as shown in Figure~\ref{fig:damagedistributions}, we set the augmentation probabilities as in Table~\ref{tab:augmentationprobabilities}. Finally, we train the model (with the selected architecture described above) using the augmented data. It takes longer training iterations to converge, and the resulted model does not perform better. This result might indicate that though individual countries may have different numbers of damages per type, the combined dataset (with three different countries) does not pose a big issue about damage type unbalancing (as shown in Figure \ref{fig:trainevalsplits}).

\begin{table}[!htb]
\centering
\caption{Augmentation probabilities for different damage types and different countries.}
\label{tab:augmentationprobabilities}
\begin{tabular}{|l|l|l|l|l|}
\hline
\textbf{Country}        & \textbf{D01} & \textbf{D10} & \textbf{D20} & \textbf{D40} \\ \hhline{|=|=|=|=|=|} 
\textit{\textbf{Czech}} & 0.2          & 0.2          & 0.0          & 0.4          \\ \hline
\textit{\textbf{India}} & 0.3          & 0.5          & 0.2          & 0.0          \\ \hline
\textit{\textbf{Japan}} & 0.0          & 0.6          & 0.3          & 0.5          \\ \hline
\end{tabular}

\end{table}

We also experiment with several other augmentation methods such as random cutout, random brightness/contrast, and cropping image augmentations, using \textit{albumentations} package \cite{info11020125}. Specifically, for cutout augmentation, we randomly set up to 8 random areas with maximum sizes of $32\times32$ with the random probability of 0.5 (i.e., there is a 0.5 chance of augmenting for every one of the eight cutout areas). For brightness/and contrasts, we experiment both with a limit of 0.3 and a random probability of 0.5. Also, for cropping, we set the min and max sizes to 512 and 540, respectively. The cropping probability used is 0.1. However, all these augmentations take a higher number of training iterations to converge, and none of them help to increase the validation accuracy.

The cutout augmentation technique does not work might due to the reason that there are already built-in dropout layers in the Detectron2 Faster R-CNN implementation. The brightness/contrast technique does not help might indicate that the training data itself already contains images with different light and weather conditions. Thus, they already have images with different brightness/contrast levels. Lastly, the cropping augmentation technique does not work might due to the reason that Faster R-CNN uses only the regions of interests of each training image (i.e., ground-truth boxes) rather than the entire image \cite{alfarrarjeh2018deep}. 

We explore also test time augmentation (TTA). Specifically, we apply flipping, resizing, and brightness/contrast augmentation techniques. We only apply horizontal flipping for flipping augmentation since the vertical flipping does not make sense in this case. For resizing, we use the following sizes (400, 500, 600, 700, 800, 900, 1000, 1100, 1200). TTA takes a longer time to make predictions because they have to make predictions for several augmented images, then combine them to produce final predictions for each image. However, neither of these approaches helps to improve prediction accuracy.

\subsection{Other hyperparameters}
Batch normalization is one of the breakthroughs in deep learning. It allows faster and more stable training by making the output distribution from one layer stable before forwarding to the next layer. This strategy also helps gradient descent by avoiding vanishing gradients. It normalizes the previous layer's output by subtracting the empirical means over the batch divided by the observed standard deviations. Therefore, there are options to change the pixel means and standard deviations over the 3 image channels (i.e., Red, Green, and Blue).

Since Detectron2 recommends not to change the standard deviations, we look into the pixel means (cfg.MODEL.PIXEL\_MEAN) from all the images in the training set as $[122.190, 122.639, 117.788]$ (in Blue, Green, and Red channels, respectively) and use them instead of the default values (generated from ImageNet dataset) as $[103.530, 116.280, 123.675]$. However, this approach does not help to enhance performance. This result might indicate that the calculated means for this dataset and those from ImageNet are not very different. Another indication would be the base model was trained with the default means and standard deviations, so changing them impacts the extracted features from the base model used.

\begin{figure}[!htb]
    \centering
    \includegraphics[width=\linewidth]{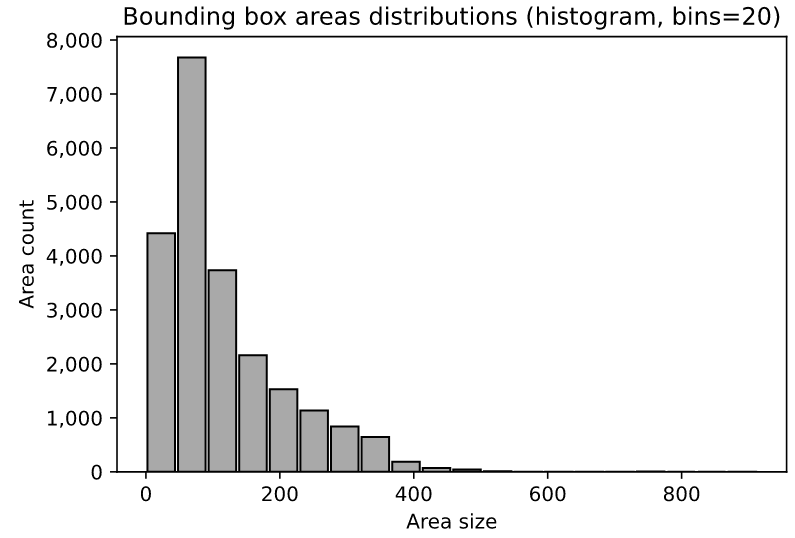}
    \caption{Bounding box area distributions using a histogram with the number of bins as 20 (range of a bin is approximately 45.6). It is observable that the areas distributed mostly around 0 to 400 squared pixels.}
    \label{fig:areadistributions}
\end{figure}

Furthermore, Faster R-CNN has a module to generate anchor boxes. These anchor boxes are generated with different sizes and ratios. Therefore, we explore our data to find appropriate box sizes and ratios. Specifically, Figure \ref{fig:areadistributions} shows the histogram of the areas of all ground-truth bounding boxes for the training set's road damages. It is observable that the areas are distributed mostly around 0 to 400 squared pixels. Therefore, we set the anchor generator sizes (cfg.MODEL.ANCHOR\_GENERATOR.SIZES) to $[[32, 64, 128]]$ instead of the default $[[32, 64, 128, 256]]$.

\begin{figure}[!htb]
    \centering
    \includegraphics[width=\linewidth]{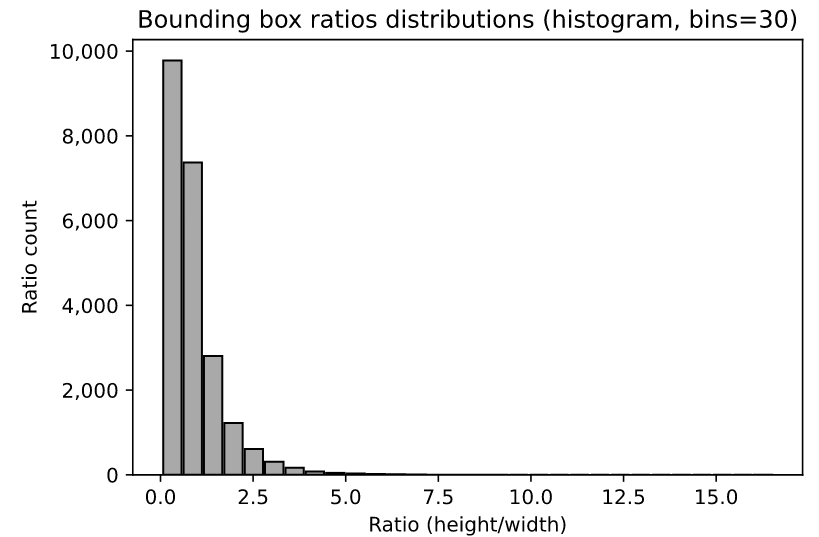}
    \caption{Bounding box ratios (height/width) distributions with the number of bins as 30 (the width of a bin is approximately 0.55). It's observable that this distribution is skewed toward the lower end.}
    \label{fig:ratiosdistributions}
\end{figure}

The ratios of the anchor boxes are calculated by their heights over their widths. Therefore, we also explore the $height/width$ distribution of all the ground-truth bounding boxes. It is observable from Figure \ref{fig:ratiosdistributions} that a high number of the ratios are distributed at the lower end. Therefore, we set the ratios (cfg.MODEL.ANCHOR\_GENERATOR.ASPECT\_RATIOS) to these values $[[0.1, 0.5, 1.0, 1.5]]$ instead of the default $[[0.5, 1.0, 2.0]]$ ratios. Training our model with these parameters does not help to improve the results. However, it helps make the learning process converges faster (around the iterations 70,000 instead of 105,000). These results indicate that specifying appropriate sizes for the anchor boxes helps the learning speed. However, the model is complicated enough to learn the boxes though the candidates are not very close to the predicted ones.

\subsection{EfficientDet model}

As discussed, the object detection research field is emerging, and related techniques evolve consistently. Specifically, Google Brain team recently published EfficientDet~\cite{tan2020efficientdet}, which claims to achieve a state-of-the-art result on Common objects in context (COCO) test-dev~\cite{lin2014microsoft}. Therefore, we also implement and train a model using EfficientDet. However, the prediction result is not as good as Faster R-CNN. Though it takes a similar amount of time to train, it does not produce a better prediction performance. This result might indicate either EfficientDet is not suitable for this dataset or need more experiments to tune its parameters for this specific case.

\section{Evaluations and suggestions}

\begin{figure}[!htb]
    \centering
    \includegraphics[width=\linewidth]{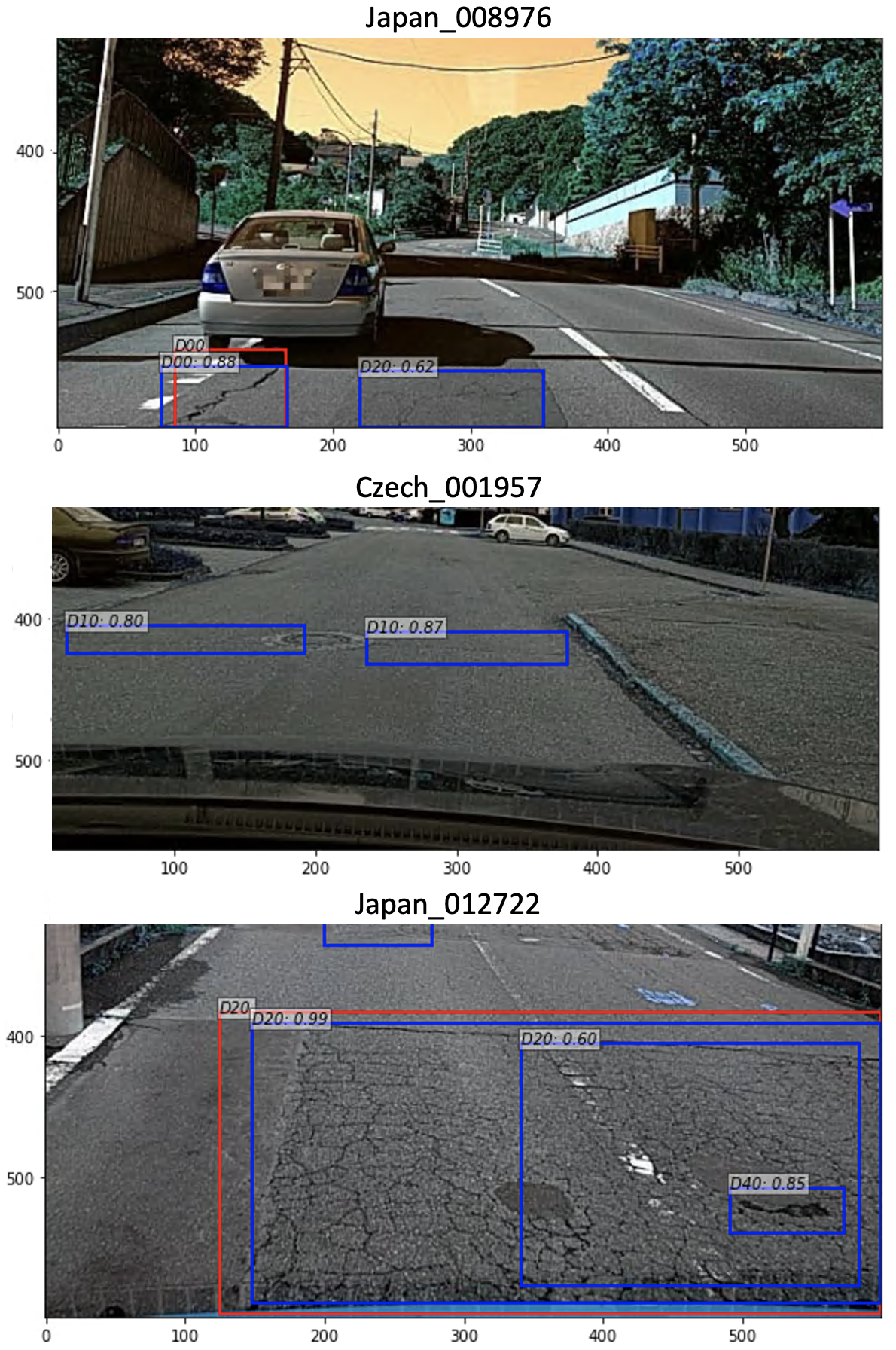}
    \caption{Examples of missing bounding boxes and labels in the training set. In this case, the red boxes are ground-truth, and the blue boxes are the predicted ones. Both have corresponding labels (for damage types), and there is also the predicted scores for the predictions.}
    \label{fig:evaluations}
\end{figure}

We also visualize the predicted bounding boxes with corresponding labels and scores to qualitatively evaluate the results. In general, predictions and ground-truth match pretty well. However, we also discover several discrepancies and found some wrong/missing ground-truth bounding boxes. Figure \ref{fig:evaluations} shows a few of these discrepancies. The red boxes are ground-truth, and the blue boxes are predicted ones. The predicted boxes also have corresponding labels and prediction probabilities (ranging from 0 to 1.0 exclusively as low to high confidence). We only show three examples in this case, and we also crop and keep only the lower parts of these pictures due to space limitations. For clarity of these pictures, we recommend interested readers to check these pictures in their original sizes from the training folder using the image file names listed on top of the pictures.

Specifically, the first picture is \textit{Japan\_008976}, while the \textit{D00} prediction matches the ground-truth, another detected damage of type \textit{D20} is obvious but is not in the annotations provided by the dataset. On the other hand, there are two horizontal cracks detected with high confidences in the picture \textit{Czech\_001957}, but there are no annotations for these in the training set. Finally, there is a pothole (\textit{D40}) damage detected in \textit{Japan\_012722} without a corresponding ground-truth. It's also worth noting from this picture (\textit{Japan\_012722}) that there are currently two overlapping bounding boxes with the same damage type (\textit{D20}), but one with 0.99 and another with 0.60 confidence scores, respectively. The predictive score threshold and the non-maximum suppression step will remove the one with lower prediction score. Thus, the predicted one bounding box for \textit{D20} matches the ground-truth one.

\begin{figure}[!tb]
    \centering
    \includegraphics[width=\linewidth]{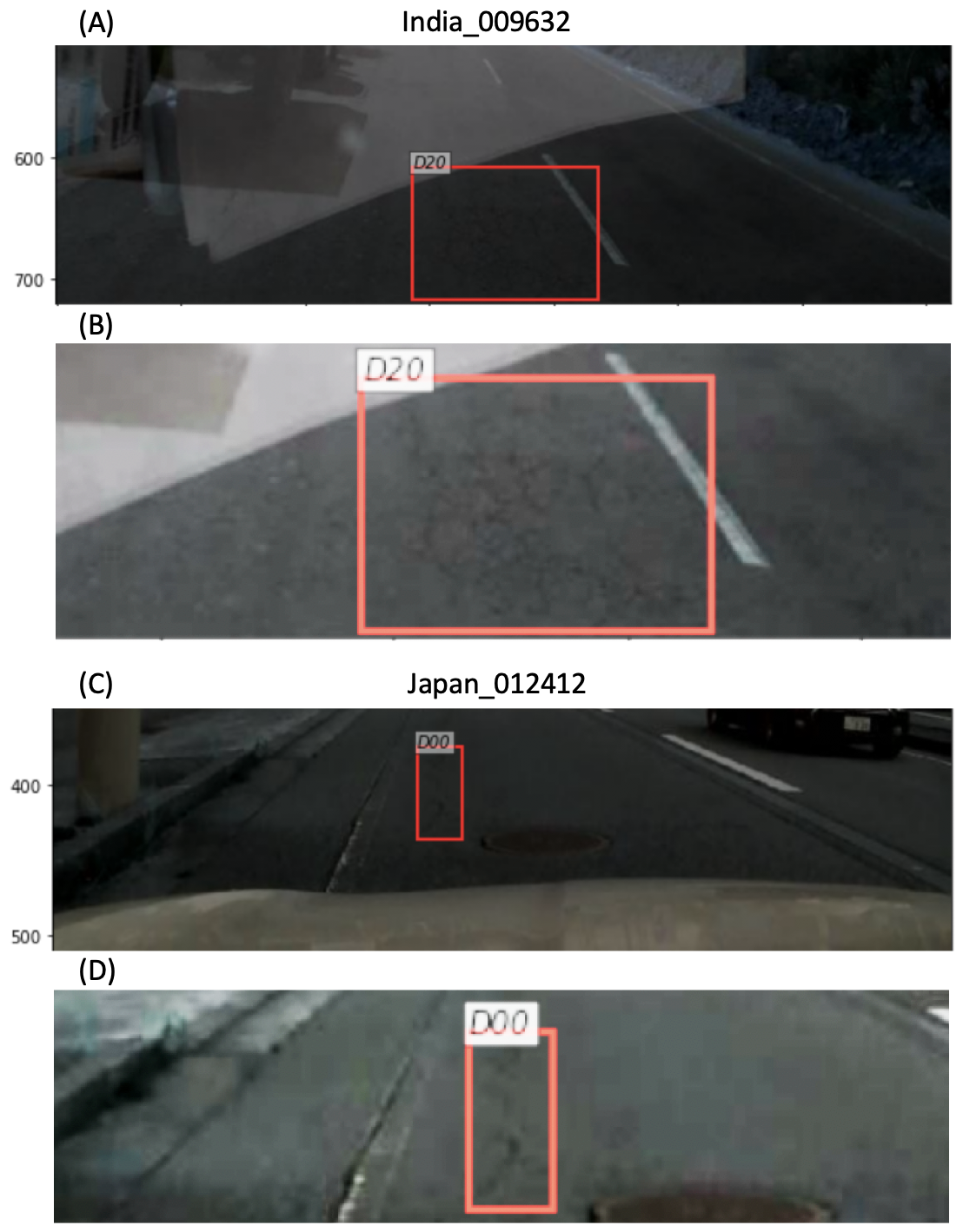}
    \caption{Examples of the case that the machine learning model outperforms humans. Panel (A) shows that our machine learning model detects a \textit{D02} type of damage in the image \textit{India\_009632}, and it is not visible to humans. It is only perceptible when we change the brightness of the picture and zoom into the region, as in panel (B). Similar story happens to the image \textit{Japan\_012412} for damage type \textit{D00} in panels (C) and (D).}
    \label{fig:brightness}
\end{figure}

In many cases, especially in darker light conditions, machine learning models can perform better than humans. For instance, Figure \ref{fig:brightness} shows the picture \textit{India\_009632} from \textit{test1} set. Our model predicts that there is a \textit{D02} type of damage in the picture (the red box in panel (A)). However, this is not clear to humans. We might think that was a wrong prediction as one can visibly check in the top panel. However, it turns out to be a correct prediction when we set the brightness of this picture to 150 (e.g., using Photoshop) and zoom into this region (200\%), as shown in panel (B). Similar  story happens to image \textit{Japan\_012412} and the detected \textit{D00} damage type in the panels (C) and (D). These examples imply that the human manual labeling approach might not be sufficient for this type of dataset. In other words, there should be a combination of both human and machine learning supports for this task.

\begin{figure}[!tb]
    \centering
    \includegraphics[width=\linewidth]{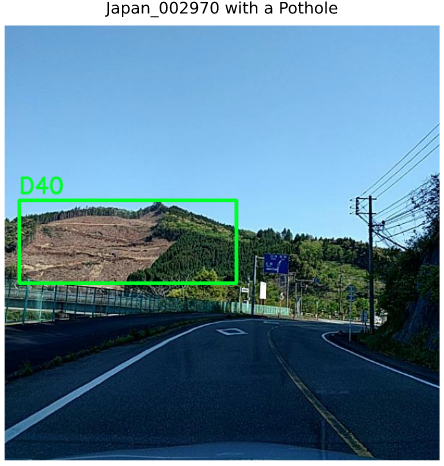}
    \caption{An image (\textit{Japan\_002970}) with wrong \textit{pothole} damage bounding box in the training set for Japan. This mistake is obvious to humans and less clear to machine learning models. Thus, it might indicate that a machine learning model generates this label, and the error was slipped through the human quality check step.}
    \label{fig:wronglabel}
\end{figure}

Deep learning needs a massive amount of labeled data, and manual labeling of road damages takes so much time and is error-prone. The reasons for errors might due to different light conditions, precisions of the bounding boxes are difficult to set, or even the confusion between damage types. Kluger et al.,~\cite{kluger2018region} suggest using a learned model to predict the bounding boxes and classify damage types of the unlabeled, collected images. These predictions then undergo the human manual checks to validate/change the bounding boxes or the labels. However, it still takes time, and there is a lot of room for errors with human manual label checks. Figure \ref{fig:wronglabel}, shows an example of such mistakes. Image \textit{Japan\_002970} (from the \textit{train} set) has a wrong \textit{pothole} damage bounding box. This wrong label is obvious to humans and less clear to machine learning models. This type of error also indicates that several labels in this current training set were generated automatically and did not get validated carefully by humans. This error would also indicate that the labeling process depends on one model learned on the initial limited amount of data.

On the other hand, in a recent study, Xie et al.,~\cite{xie2020self} claim to have state-of-the-art ImageNet classification results using self-learning with noisy student technique. We suggest using this approach with the following steps. First, train a model on existing images with labeled bounding boxes and classes. Use it as a teacher to generate annotations for other unlabeled images. Combine the newly predicted annotations (with higher confidences) with the existing annotations, then train a larger model called a student model. Use this learned student as a teacher and repeat the process. This approach incrementally increases the labeled data instead of training on a single model then uses that single model to predict labels as in the previous approach. 

It would always be useful to finally pass the pseudo, predicted labels through another human inspection to validate or adjust the bounding boxes and labels as a quality check. However, it's worth noting that one should also change the brightness and zoom into the predicted regions for better quality checks if they are not apparent to humans. These cases can be done manually or detected (using overall light conditions) automatically adjust the pictures to help the validation process.

\section{Selected model, results, and limitations}
Finally, we selected Faster R-CNN with X101-FPN as the architecture to tackle the tasks of this competition. This approach (with the parameters described above) results in F1 scores of 51.0\% and 51.4\% for the \textit{test1} and \textit{test2} sets of the challenge, respectively. The low F1 score is arguably acceptable due to several issues with the ground-truth in the training sets, as detailed in the previous section. However, the main limitation of using Faster R-CNN with X101-FPN is that it is slower to train and has a longer prediction time than other model types such as YOLO and SSD.

\section{Conclusion}
This work explores different state-of-the-art object detection methods and their applicability for road damage detection and classification tasks. Specifically, we experiment with Detectron2's Faster R-CNN implementation with different base models and configurations using the Global Road Damage Detection Challenge 2020 dataset. We also examine other state-of-the-art object detection methods and various techniques like training time, testing time augmentations, context information, and post-processing. However, these methods are not suitable for the damaged road objects, and their effects are not satisfactory. In other words, the results indicate that Faster R-CNN with X101-FPN base model and Detectron2's default configurations produce good prediction results for these tasks (F1 score of approximately 51.0\% for both test sets) and is also general enough to be used in different countries. We also visualize and qualitatively evaluate the existing labeling quality and suggest using the noisy student approach to improve the road damage labeling process. 

The source codes of the experiments are available at the Github page of this project: \url{https://github.com/iDataVisualizationLab/roaddamagedetector.}

\bibliographystyle{./IEEEtran}
\bibliography{./IEEEabrv,./references}

\end{document}